\documentclass{article}
\usepackage{xcolor}
\usepackage{algorithm}
\usepackage[noend]{algpseudocode}
\usepackage{adjustbox}
\usepackage{spconf,amsmath,graphicx}
\usepackage{nccmath}
\usepackage{url}
\usepackage{amsthm}
\PassOptionsToPackage{hyphens}{url}\usepackage{hyperref}
\usepackage{mathtools}
\DeclarePairedDelimiter\abs{\lvert}{\rvert}%
\newcommand{\ignore}[1]{}
\newcommand{\printfnsymbol}[1]{%
  \textsuperscript{\@fnsymbol{#1}}%
}
\title{Semi supervised segmentation and graph-based tracking of 3D nuclei in time-lapse microscopy }
\name{S. Shailja$^{*}$\thanks{* denotes equal contribution}, Jiaxiang Jiang$^{*}$, and B.S. Manjunath  }
\address{ Department of Electrical and Computer Engineering,\\ University of California Santa Barbara, CA, United States}
\begin{document}
%
\maketitle
\begin{abstract}
We propose a novel weakly supervised method to improve the boundary of the 3D segmented nuclei utilizing an over-segmented image. This is motivated by the observation that current state-of-the-art deep learning methods do not result in accurate boundaries when the training data is weakly annotated. Towards this, a 3D U-Net is trained to get the centroid of the nuclei and integrated with a simple linear iterative clustering (SLIC) supervoxel algorithm that provides better adherence to cluster boundaries. To track these segmented nuclei, our algorithm utilizes the relative nuclei location depicting the processes of nuclei division and apoptosis. The proposed algorithmic pipeline achieves better segmentation performance compared to the state-of-the-art method in Cell Tracking Challenge (CTC) 2019 and comparable performance to state-of-the-art methods in IEEE ISBI CTC2020 while utilizing very few pixel-wise annotated data. Detailed experimental results are provided, and the source code is available on GitHub \footnote{\url{https://github.com/s-shailja/ucsb_ctc}}.
\end{abstract}
\section{Introduction}
\label{sec:intro}
Nuclei migration and proliferation are two important processes in tissue development at early embryonic stages. Optical time-lapse microscopy is the most appropriate imaging modality to visualize these processes. Such microscopy recordings can generate massive data, allowing for a detailed analysis of nuclei physiology and properties. To gain biological insights into nuclei behavior, it is often necessary to identify individual nucleus (segmentation) and follow them over time (tracking). However, manual data analysis is infeasible due to the large amount of data acquired. Also, segmenting the nuclei in the microscopic images is a daunting task because of the presence of noise that affects their visual appearance as well as shape.
    
Convolutional neural networks (CNNs), especially U-Net \cite{unet} have been widely used in the cell and nuclei segmentation context because of their superior segmentation performance. The general nuclei segmentation problem can be formulated as either instance segmentation or semantic segmentation. The instance segmentation~\cite{instance1, instance2, instance3} tends to give good detection accuracy but not segmentation accuracy while semantic segmentation~\cite{semantic1, semantic2, semantic4} is viable but loses accuracy around the border of nuclei when training data is sparsely annotated. Along this line, researchers have proposed an active learning tool to improve nuclei boundaries but it is dependent on user feedback~\cite{delibaltov2016cellect}. Based on the segmentation or detection of nuclei, most successful nuclei tracking methods rely on the Viterbi algorithm~\cite{magnusson2014global}. These techniques construct tracking as a global optimization problem and utilize absolute nuclei location to detect their trajectories. Therefore, it is not efficient. 

In this paper, we propose a novel semi supervised nuclei segmentation method utilizing Simple linear Iterative Clustering (SLIC) boundary adherence and a graph-based tracking algorithm utilizing relative cell location information. The main contributions of this paper are two-fold. First, we propose a novel method to improve nuclei boundary detection and thereby segmentation for quantitative cell morphology. Second, a novel graph-based tracking method that leverages the stable relative nuclei location in consecutive video frames is developed. We also show that the algorithm can be extended to other datasets giving comparable results.
\section{Segmentation algorithm} 
\begin{figure}[htb]
\scriptsize
            \centering
            \centerline{\includegraphics[scale=0.36]{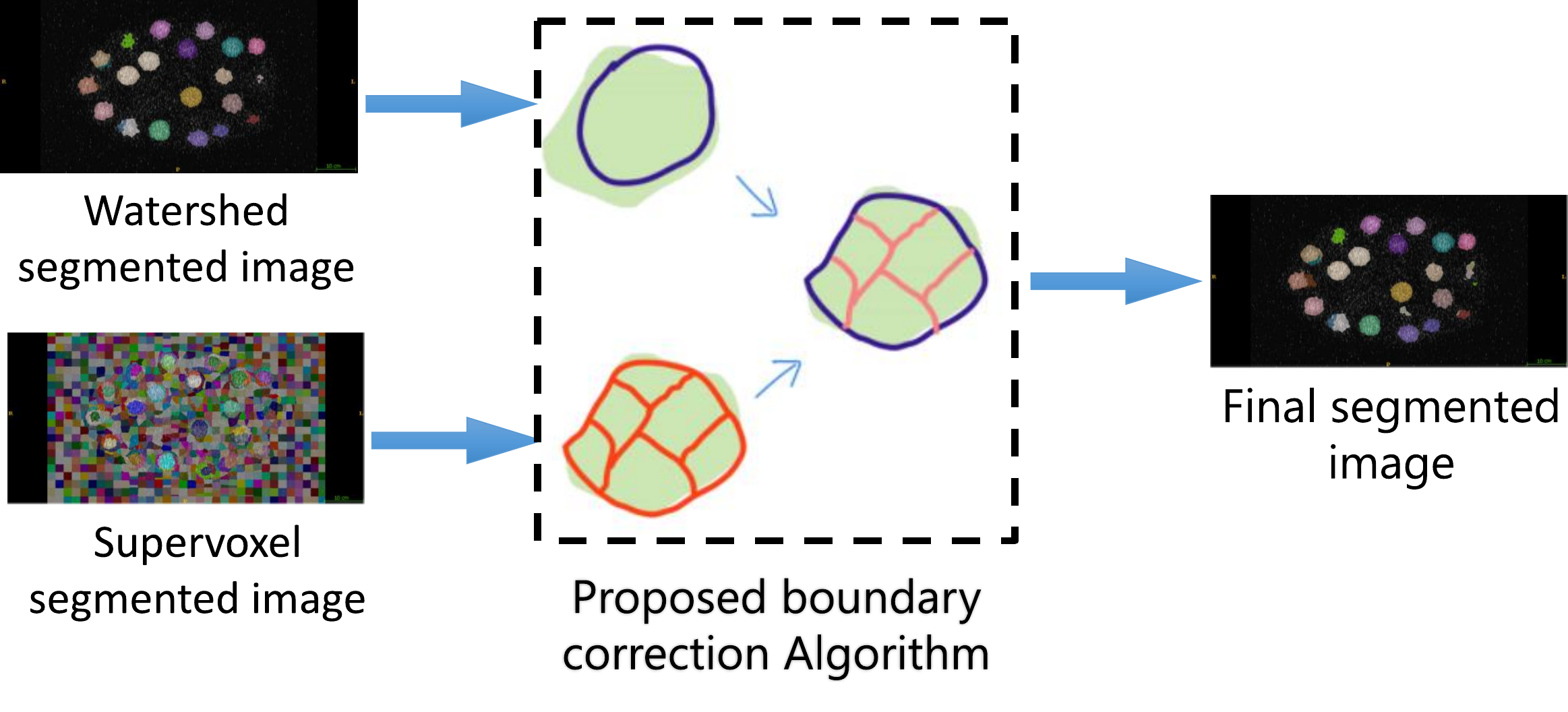}}
        \caption{\footnotesize Schematic diagram of the proposed segmentation method.}
        \label{fig:seg_schematic}

    \end{figure}

This section describes the details of the proposed segmentation method for weakly annotated data that takes advantage of the boundary correction algorithm using supervoxels. 
In time-lapse videos of three dimensional image stacks, annotating data points is a time and resource consuming process. As a result, only a few 2D slices of expert annotated ground truth is available. This is not sufficient to train a deep learning model for accurate semantic segmentation. Besides, learning models only based on such sparse reference data points are not scalable. This emphasizes the need of a semi-supervised algorithm for accurate segmentation. Moreover, most of the methods are texture-based and not accurate for the boundaries of the nuclei which play an important role in understanding nuclei morphology and proliferation. Towards that end, we present a robust and scalable method to segment and track nuclei. Our algorithm utilizes over-segmented image stacks to improve the segmentation of the given image by mainly correcting the boundaries of a texture-based method. In order to do so, we first segment the image stack using standard 3D watershed segmentation. In parallel, we also obtain the over-segmented image stack using supervoxel segmentation. Finally, we propose our boundary correction algorithm that takes advantage of both segmentation methods to improve the boundaries. The schematic of the proposed segmentation algorithm is shown in the Figure~\ref{fig:seg_schematic}.
    
To get an approximate segmentation, we utilize the 3D watershed segmentation. Watershed segmentation algorithm~\cite{vincent1991watersheds} is applied to the 3D probability map of nuclei that is generated by a very small CNN consisting of 3D convolution layers. This gives a rough segmentation of the image but generates labels for the nuclei.
We denote the output segmented image from the watershed as $I^{w}_{\text{seg}}$ with $C^{w}_{k}$ representing the set of voxels that form the cluster for nucleus $k$. 

In order to get the over-segmented image, supervoxel segmentation method is used. SLIC~\cite{achanta2012slic} is a method for generating supervoxels from images using an adaptation of k-means clustering that uses a distance function with both intensity and distance similarity terms. We denote the output over-segmented image as $I^{s}_{\text{seg}}$ with $C^{s}_{k}$ representing the set of voxels that form the cluster for nucleus $k$.

\subsection{Boundary correction algorithm}
We propose a novel boundary correction algorithm that takes advantage of the over-segmented supervoxels in $I^{s}_{\text{seg}}$ (from SLIC) to improve watershed segmented image $I^{w}_{\text{seg}}$. We know that $I^{s}_{\text{seg}}$ is the over-segmented image that increases the chances that nuclei boundaries are extracted at the cost of creating many false boundaries within the nuclei. Hence, we have, 
\begin{equation*}
    \abs{C^{s}} > \abs{C^{w}}, \quad C^{s} = \{C^{s}_{1}, C^{s}_{2}, C^{s}_{3},...\}
\end{equation*}
where $\abs{\cdot}$ denotes the total number of clusters in the segmented image. Consider $C^{s}_{i} \in C^{s}$ and $C_j^w \in C^w$ for all clusters $i,j \in \{1,2,3,...\} \times \{1,2,3,...\}$. Define a cluster correlation parameter $K_{ij}$ as 
\begin{equation}
    K_{ij} \overset{\Delta}{=} \abs{C^s_i \cap C^w_j}.
\end{equation}
Then, 
we compute $C^{s}_{j^*}$ for each label $i$, where
\begin{equation}
  j^* = \arg \max_{j} K_{ij}.
\end{equation}

For every $i$, we obtain $j^*$ that maximizes the clustering correlation between the clusters of the watershed segmented image and the SLIC segmented image. This ensures that for each watershed segmented nuclei, we have an accurate boundary of the nuclei supervoxels. Hence, we get improved nuclei boundaries in the final segmented image $S$ with nuclei represented by $C^{s}_{j^*}$. Supervoxels based boundary correction algorithm refines the watershed segmented boundaries in $I^{s}_{\text{seg}}$ by taking into account the boundaries of the supervoxels. Boundary adherence is one of the important properties of SLIC supervoxel as it reflects how supervoxel boundaries fit the nuclei borders.

\begin{algorithm}[b]
  \scriptsize
   \caption{Cell Nuclei Adjacency Graph Construction}
    \begin{algorithmic}
      \Function{Adjacency}{Segmented Image $S$}
        \State Initiate AdjacentGraph A
        \State Initiate drawboard (same shape as $S$) with 0 entries
        \For{$i = 1$ to number of Nuclei}
            \For{$j = 1$ to number of Nuclei}
            \If{the voxel belong to Nucleus i or j}
            \State entry of drawboard is set to 1
            \EndIf
            \If{i not equal to j}
                \State 3D dilation of drawboard
                \State Connected component analysis
                \If {number of component is 1}
                    \State append j to ith entry of AdjacentGraph
                \EndIf
             \EndIf
             \EndFor
         \EndFor
      \EndFunction
\end{algorithmic}\label{algorithm:1}
\end{algorithm}
\vspace{-6pt}
\section{Graph-based tracking}
\vspace{-6pt}

\begin{figure}[b]

            \centering
            \centerline{\includegraphics[width=5cm]{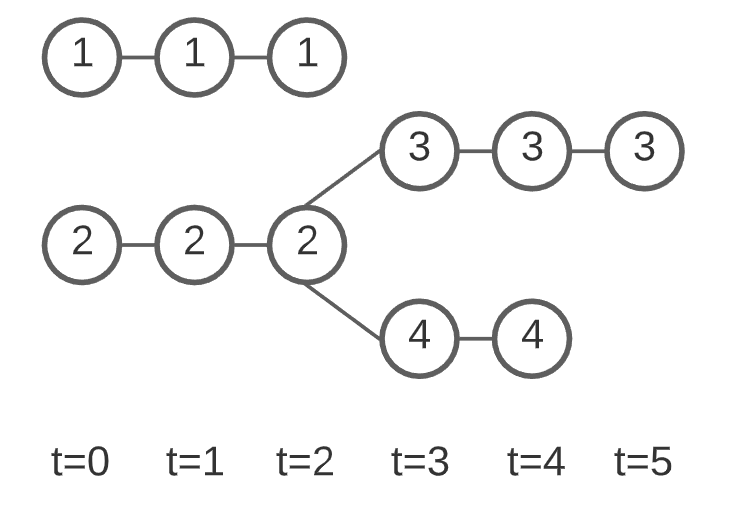}}

        \caption{\footnotesize Nucleus tracking illustration: number is used to represent the unique ID for each track of nucleus. At t=3, nucleus 2 divides into 2 new nuclei 3 and 4}
        \label{fig:tracking}

\end{figure}
Nuclei tracking is to reconstruct the lineage of cell nuclei and match related cell nuclei across the whole video sequence. The tracking process will give a trajectory for each individual nucleus as shown in Figure~\ref{fig:tracking}. In the traditional Viterbi cell tracking algorithm, its complexity is $\mathcal{O}(TM^{4})$ where $T$ is the length of the video sequence and $M$ is the maximum number of nuclei. This is because the complexity of the Viterbi algorithm is linear in $T$, there can be $N^2$ pairs of nuclei in any two frames, and every such pair can have as many swap arcs between them as there are pre-existing tracks. A general assumption, as shown in~\cite{trackviterbi}, is that only certain nuclei events (apoptosis, division, etc) can happen and thus reduces possible swap arcs to some constant. This reduces the whole tracking complexity to be quadratic in the number of nuclei. We propose an adjacency graph-based nuclei tracking algorithm utilizing nuclei relative location information to reduce the complexity to $\mathcal{O}(TM^{2})$ without any assumption of nuclei events while achieving comparable results.

    
A nuclei adjacency graph $G(V,E)$ is an undirected weighted graph built based on the segmented image $S$. In $G(V,E)$, 
each vertex $v_i \in V$ represents each individual nucleus. 
For each pair of vertices $(v_i,v_j)$, there is an edge $e_i \in E$ connecting them. 
The weight $w_i \in W$ of the edge $e_i$ is the minimum distance between nucleus $i$ and $j$. 
The minimum distance is computed as number of times morphology dilation operations of nucleus $i$ and $j$ need to be applied until nucleus $i$ and $j$ become a single component. The step-by-step details of the method is described in Algorithm~\ref{algorithm:1}. Now, the natural solution to the tracking problem is defined as finding the similar vertices in two graphs. Therefore, we build a feature $\textbf{f}_{\text{track}}^i$ vector for each vertex $v_i$.  $\textbf{f}_{\text{track}}^i$ is a three dimensional feature vector with entries $vol^i$ and $\textbf{f}_{\text{loc}}^i$ . $vol^i$ is a scalar representing the volume of each nucleus. $\textbf{f}^i_{\text{loc}}$ is a two dimensional vector $(N^i,D^i)$, where $N^i$ is the total number of neighbor nuclei and $D^i$ is the average distance from all other nuclei. 
Given the adjacency graph $G(V,E)$ of the segmented image stack, for each vertex $i$ in the graph, the location feature vector can be expressed as\begin{equation}
    \textbf{f}_{\text{loc}}^i=(N^i,D^i)=(deg(v_i),wdeg(v_i))
\end{equation}
where $v_i \in V$, $deg(v_i)$ is the cardinality of $N^i$, and $wdeg(v_i)$ is the weighted degree of the vertex $v_i$ defined as
\begin{equation}
    wdeg(v_i)=\frac{\sum_{j} w_{ij}}{degree(v_i)}
\end{equation}
where $degree(v_i)$ represents the degree of the vertex $v_i$. 
The procedure of constructing $\textbf{f}_{\text{track}}^i$ is described in Algorithm ~\ref{Algorithm:2}.

 \begin{algorithm}[t]
 \scriptsize
   \caption{Tracking Feature Computation}
    \begin{algorithmic}
    
      \Function{TrackFeature}{Segmentation and G(V,E)}
        \For{$i = 1$ to number of nuclei}
            \State Initiate Feature Vectors $\textbf{f}_{\text{track}}^i$
        \EndFor
        \For{$i = 1$ to number of nuclei}
            \State cell volume $vol^i$ = number of voxel inside nuclei i $\times$ voxel resolution
            \State $wdeg(v_i)=\frac{\sum_{j} w_{ij}}{degree(v_i)}$
            \State  $\textbf{f}_{\text{track}}^i$ =$(vol^i_i,deg(v_i),wdeg(v_i))$ 
        \EndFor
       \EndFunction
\end{algorithmic}\label{Algorithm:2}
\end{algorithm}

After computing $\textbf{f}_{\text{track}}^i$ for all nodes in two consecutive frames, we link two nodes from different frames based on the following similarity measurement $sim$ defined as
\begin{equation}
\begin{aligned}
    sim = \frac{|S_{1i}-S_{2j}|}{S_{1i}}+\frac{|deg_1(v_i)-deg_2(v_j)|}{deg_1(v_1)}+\\
    \begin{aligned}
    \frac{|wdeg_1(v_1)-wdeg_2(v_2)|}{wdeg_1(v_1)}
    \end{aligned}
\end{aligned}
\end{equation}
where $i$ and $j$ denote two nodes from different frames. We define $sim$ so that we can allow different units of entries in $\textbf{f}_{\text{track}}^i$. We find $i^{*}$ and $j^{*}$ that minimizes $sim$. $i^{*}$ and $j^{*}$ are linked only when their $sim$ is below a set threshold value. If some nucleus in the previous frame has no linked nucleus in the latter frame, it means apoptosis happens or the nucleus leaves the field of view. If some nucleus in the latter frame has no linked nucleus in the previous frame, it means the new nucleus comes into the field of view. Thus, based on this measure, we can track each nucleus in consecutive frames of the recordings. The complexity for finding all possible links in consecutive frames is $\mathcal{O}(M^{2})$. Therefore, the whole tracking complexity is $\mathcal{O}(TM^{2})$.
\section {Experimental Results}
\vspace{-6pt}
\subsection{Dataset}
\begin{figure}[h!]
\scriptsize
            \centerline{\includegraphics[scale = 0.58]{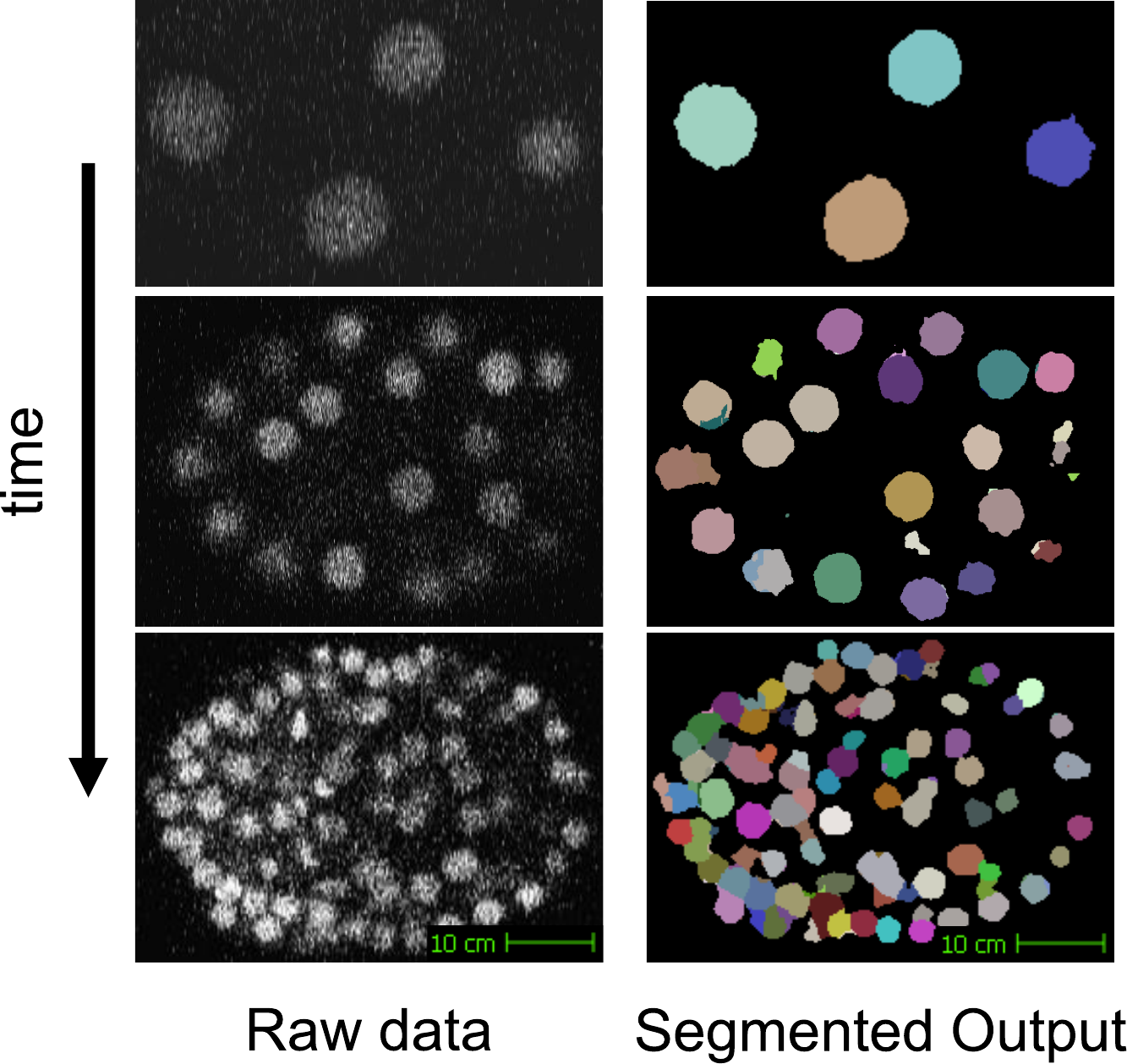}}
        \caption{\footnotesize 2D slices of the raw and segmented data for 3D N3DH-CE at different time instances in minutes.}
        \label{fig:raw}

    \end{figure}

\vspace{-6pt}
The time series dataset in CTC2020 consists of 3D time-lapse video sequences of fluorescent counterstained nuclei microscopy image of \textit{C.elegans} developing embryo as shown in Figure~\ref{fig:raw}. Each voxel size is $0.09 \times 0.09 \times 1.0 $ in microns. Time points were collected once per minute for $(5-6)$ hours. There are 2 videos in the training set and 2 videos in the challenge (testing) dataset. The details of the data is summarized in the Table~\ref{tab:dataset}. Gold-standard corpus containing human-origin reference annotations are referred as gold truth (GT). We used the annotated GT from training dataset for our training and validation procedure. The GT labels are withheld for the test videos and used to evaluate the model. 

 \begin{table}[ht]
\scriptsize
    \centering
    
    \begin{tabular}{|c|c|c|c|c|c|}
        \hline
    Dataset & Video & Dimension & \#frames & \#seg GT & \#track GT\\ \hline
        
        Training & 01 & $512\times708\times35$ & 250 & 5& 195\\
        & 02 & $512\times712\times31$ & 250 & 5 & 190 \\ 
        Testing & 01 & $512\times712\times31$ & 190  & - & -\\ 
         & 02 &$512\times712\times31$ & 140 & - & - \\ \hline
    \end{tabular}
    \caption{\scriptsize Dataset description for CTC2020.  \# denotes the number of, for example, \#frames refers to the number of frames in the video}
    \label{tab:dataset}
\end{table}
\vspace{-10pt}

\subsection{Evaluation Metrics}  \label{metrics}
\vspace{-6pt}
Five evaluation metrics, detection accuracy (DET), segmentation accuracy (SEG), tracking accuracy (TRA), Cell Segmentation Benchmark ($\text{OP}_\text{CSB}$), and Cell Tracking Benchmark ($\text{OP}_\text{CTB}$) are commonly used in the nuclei segmentation and tracking problem. DET is used to quantify how well each given nuclei has been identified. It is defined based on Acyclic Oriented Graph Matching (AOGM-D)~\cite{matula2015cell} measure for detection as 
\begin{equation*}
   \text{DET} = 1 - \frac{\min(\text{AOGM-D}, \text{AOGM-D}_0)}{\text{AOGM-D}_0}
\end{equation*}
where AOGM-D is the cost of transforming a set of nodes provided by the algorithm into the set of ground truth nodes; AOGM-D$_0$ is the cost of creating the set of ground truth nodes from scratch. DET ranges between 0 to 1 (1 means perfect matching). SEG is a statistic used to measure the similarity of the segmented nuclei and ground-truth nuclei. It is defined based on the Jaccard similarity index (J) as
\begin{equation*}
    \text{SEG} = \frac{|R\cap S|}{|R\cup S|}
\end{equation*}
where $R$ and $S$ denotes the set of pixels in the ground-truth and prediction, respectively. SEG ranges between 0 to 1 (1 means perfect matching). TRA measures how accurately each nuclei has been identified and followed in successive frames of the video. It is defined based on AOGM as
\begin{equation*}
    \text{TRA} = 1 - \frac{\min(\text{AOGM}, \text{AOGM}_0)}{\text{AOGM}_0}
\end{equation*}
where AOGM$_0$ is the AOGM value required for creating the reference graph from scratch. SEG ranges between 0 to 1 (1 means perfect tracking). For direct comparison of the methods, Cell Segmentation and Tracking Benchmark is evaluated using $
\text{OP}_\text{CSB}$ and $\text{OP}_\text{CTB}$, defined as
\begin{equation*}
    \text{OP}_\text{CSB} = \frac{\text{DET} + \text{SEG}}{2}, \quad 
    \text{OP}_\text{CTB} = \frac{\text{SEG} + \text{TRA}}{2}.
\end{equation*}
\vspace{-20pt}
\subsection{Segmentation Performance}
\vspace{-6pt}
The results of our proposed segmentation algorithm is shown in Figure~\ref{fig:raw}. We evaluated the method using the scheme proposed in the online version of the Cell Tracking Challenge. Results of our proposed method on the CTC2020 test set is shown in the Table~\ref{tab:seg_results_2020}. This demonstrates that our algorithm outperforms the benchmark method for nuclei segmentation for CTC2019 (MPI-GE). It is also comparable to the state-of-art method for CTC2020 (KIT-Sch-GE) while utilizing a very few number of annotated training data. We can assume that the ground truth annotations for testing dataset is low in number and weakly annotated similar to the data made available to us. So, the minor difference in the algorithmic performance based on the evaluation metric used is not significant. We present fairly robust 3D segmentation method that generalizes across datasets (as discussed in the next section) using about 100 times less annotated data points than \cite{lofflerkit} and achieve competitive performance.
\begin{table}[htbp]
\scriptsize
    \centering
   
    \begin{tabular}{|c|c|c|c|c|c|}
        \hline
    Method & DET & SEG  & OP$_{\text{CSB}}$ & \#Pixel-wise annotation used \\ \hline
       
        KIT-Sch-GE \cite{lofflerkit} & 0.915 & 0.729  & 0.830  &997 \\ 
        \textbf{UCSB-US \cite{shailjaucsb}} & \textbf{0.927} & \textbf{0.705}  &\textbf{ 0.816} &\textbf{10} \\
        MPI-GE \cite{mpige} & 0.930 & 0.688  & 0.801 &10\\ \hline
    \end{tabular}
 \caption{\scriptsize Cell Segmentation Benchmark for N3DH-CE dataset in IEEE ISBI CTC2020. Our proposed algorithm (UCSB-US) was placed second in the CTC2020. DET, SEG and ($\text{OP}_\text{CSB}$) are defined in Section~\ref{metrics}. }
    \label{tab:seg_results_2020}
    
\end{table}

\vspace{-15pt}
\subsection{Tracking Performance}
\vspace{-6pt}
In Table~\ref{tab:tra_results_2020}, we compare the nuclei tracking performance of the proposed method with the state-of-the-art methods defined in Section \ref{metrics}. 
Both KIT-SCH-GE and KTH-SE optimize the global tracking process while our methods utilize the relative nuclei location information to do the local optimization.
Our local optimized tracking process is more efficient and still achieves the comparable results (difference is less than 1\%) as shown in the Table ~\ref{tab:tra_results_2020}.
\begin{table}[htbp]
\scriptsize
    \centering
    
    \begin{tabular}{|c|c|c|c|c|c|}
        \hline
    Method & SEG & TRA  & OP$_{\text{CTB}}$ \\ \hline
       
        KIT-Sch-GE \cite{lofflerkit}& 0.729 & 0.886  & 0.808  \\ 
         KTH-SE \cite{magnusson2014global} & 0.662 & 0.945  & 0.803 \\
        \textbf{UCSB-US\cite{shailjaucsb}} & \textbf{0.705}   &\textbf{ 0.895} &\textbf{0.800} \\ \hline
    \end{tabular}
\caption{\scriptsize Cell Tracking Benchmark for N3DH-CE dataset in IEEE ISBI CTC2020. Results of our proposed algorithm (UCSB-US) was placed third in the CTC2020. SEG, TRA and ($\text{OP}_\text{CTB}$) are defined in Section \ref{metrics}}.
    \label{tab:tra_results_2020}
    \vspace{-15pt}
\end{table}

To demonstrate that our algorithmic pipeline can be easily extended, we experimented with another dataset from CTC. We used infected C3DL-MDA231 human breast carcinoma cells for this purpose. Without fine tuning the hyperparameters to this specific dataset, we achieve comparable results to the state-of-the-art method with DET = 0.839, SEG = 0.545, TRA = 0.795, $OP_{CSB}$ = 0.692, and $OP_{CTB}$ = 0.670. The source code is publicly available on GitHub \footnote{\url{https://github.com/s-shailja/ucsb_ctc}}.

\section{Discussion \& Conclusion}
In this paper, we have presented a novel weakly supervised 3D nuclei segmentation method that consists of deep learning based nuclei detection, watershed segmentation, and a boundary correction algorithm using supervoxels. Specifically, we demonstrate that the proposed segmentation method explicitly carries boundary information of the nuclei thus improving performance of the traditional watershed segmentation. We also show that our resource efficient algorithm exploits partially labeled data to achieve competitive performance. 

Additionally, we present a simple and efficient graph-based tracking algorithm utilizing relative nuclei location information extracted from the adjacency graph. The widely used Viterbi algorithm models the whole sequence of detected cells in video as a directed acyclic graph to solve a global optimization problem. In contrast to this, our frame-by-frame tracking algorithm does not require an entire recorded sequence and can be applied in real time applications while still maintaining comparable results to state-of-the-art methods. For future work, we intend to further evaluate the reproducibility of our approach on additional datasets. We also intend to study a joint optimization problem which includes segmentation as well as tracking. This will explore how tracking could be used as feedback to improve segmentation.

\section{Acknowledgments}
We would like to thank Angela Zhang for helpful discussions and Dr. Robby Nadler for proofreading the manuscript. This research is supported by a National Science Foundation (NSF) award \# 1664172.

\bibliographystyle{IEEEbib}
\bibliography{strings,refs}

\end{document}